\documentclass{article}

\usepackage{arxiv}

\makeatletter
\renewcommand{\@toptitlebar}{\hrule height 2\p@ \vskip 0.14in \vskip -\parskip}
\renewcommand{\@bottomtitlebar}{\vskip 0.16in \vskip -\parskip \hrule height 2\p@ \vskip 0.05in}
\makeatother

\usepackage[utf8]{inputenc}
\usepackage[T1]{fontenc}
\usepackage{hyperref}
\usepackage{url}
\usepackage{booktabs}
\usepackage{amsmath}
\usepackage{amsfonts}
\usepackage{amssymb}
\usepackage{nicefrac}
\usepackage[expansion=false]{microtype}
\usepackage[final]{graphicx}
\setkeys{Gin}{draft=false}
\usepackage{xcolor}
\usepackage{caption}
\usepackage{subcaption}
\usepackage{placeins}
\usepackage[numbers]{natbib}
\usepackage{doi}

\definecolor{panelrule}{HTML}{B9C0BA}
\newlength{\hdrw}

\newcommand{\hdrsub}[1]{\\[-1pt]{\footnotesize #1}}

\newlength{\cellw}
\newcommand{\colhdr}[1]{%
  \parbox[b]{\cellw}{\centering\small #1\strut\\[1.5pt]%
    {\color{panelrule}\rule{\cellw}{0.4pt}}}}
\newcommand{\colhdrours}[1]{%
  \parbox[b]{\cellw}{\centering\small\bfseries\boldmath\color{hidark}#1\strut\\[1.5pt]%
    {\color{hitheme}\rule{\cellw}{2pt}}}}
\newcommand{\colimg}[1]{{%
  \setlength{\fboxsep}{0pt}\setlength{\fboxrule}{0.7pt}%
  \fcolorbox{panelrule}{panelrule}{#1}}}
\newcommand{\colimgours}[1]{{%
  \setlength{\fboxsep}{0pt}\setlength{\fboxrule}{0.7pt}%
  \fcolorbox{hiaccent}{hiaccent}{#1}}}
\newcommand{\gridsetup}[1]{\setlength{\cellw}{#1}\setlength{\tabcolsep}{2pt}%
  \renewcommand{\arraystretch}{1}}

\newlength{\rowh}

\definecolor{hitheme}{HTML}{95FC65}
\definecolor{hiaccent}{HTML}{95FC65}
\definecolor{hibright}{HTML}{95FC65}
\definecolor{hitint}{HTML}{EFFEE8}
\definecolor{hidark}{HTML}{32810E}
\definecolor{hideep}{HTML}{20520A}
\definecolor{hilogo}{HTML}{00FF00}

\newcommand{\res}[1]{$#1^{3}$}
\newcommand{\Hi}{Hi3D~3.0}
\newcommand{\Tw}{Twinkle3D}

\title{\ensuremath{\vcenter{\hbox{\includegraphics[height=2.6em]{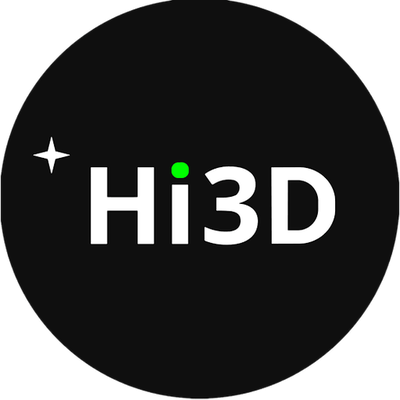}}}}%
  \hspace{0.55em}%
  \ensuremath{\vcenter{\hbox{\begin{tabular}{@{}c@{}}Hi3D 3.0 (Twinkle3D): Object-specific \\ 3D Asset Generation with high resolution\end{tabular}}}}}

\author{
  Hi3D Team \\
  \texttt{\href{https://www.hi3d.ai/}{hi3d.ai}}
}

\date{}

\renewcommand{\headeright}{Technical Report}
\renewcommand{\undertitle}{Technical Report}
\renewcommand{\shorttitle}{Hi3D 3.0 (Twinkle3D)}

\hypersetup{
  colorlinks=true,
  linkcolor=hidark,
  citecolor=hidark,
  urlcolor=hidark,
  bookmarksnumbered=true,
  pdftitle={Hi3D 3.0 (Twinkle3D): Object-specific 3D Asset Generation with high resolution},
  pdfsubject={cs.CV, cs.GR},
  pdfauthor={Hi3D Team},
  pdfkeywords={image-to-3D, object-specific fidelity, geometry generation, 3D VAE, high-resolution generation, Twinkle3D},
}

\begin{document}
\maketitle
\vspace{-0.55in}

\begin{center}
  {\itshape A geometric twin is more than a look-alike.}\\[0.25em]
  {\small \textbf{Product}\ \url{https://www.hi3d.ai/}\quad
   \textbf{Project page}\ \url{https://mathmagic-official.github.io/twinkle3d/}}
\end{center}
\vspace{0.1em}
\noindent\begin{minipage}{\linewidth}
  \centering
  \includegraphics[width=1.000\linewidth]{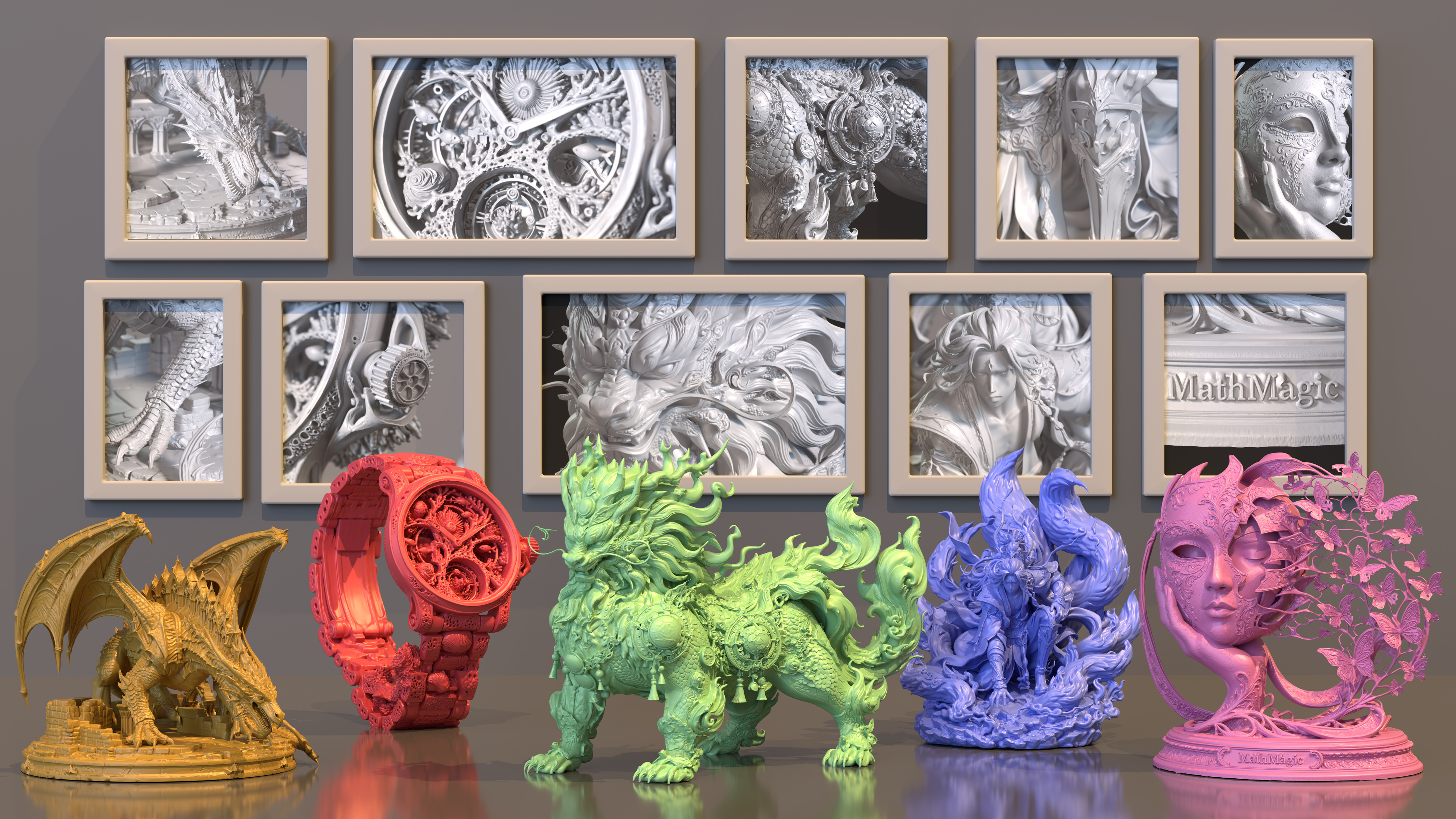}
  \captionof{figure}{\textbf{Meshes generated by \Hi{} from single images.} Framed
  on the wall: clay-shaded close-ups of the same meshes.}
  \label{fig:teaser}
\end{minipage}
\vspace{0.1em}

\begin{abstract}
Image-to-3D generation has become increasingly capable of producing objects that closely resemble the input image, and an outstanding challenge is to reproduce the depicted object itself, including the specific geometry that defines it. Inscriptions, brand marks, and repeated structures are frequently distorted or lost, despite being critical to object identity. We present Hi3D 3.0, an image-to-3D generation system targeting object-specific fidelity, with Twinkle3D as its geometry model for generating watertight triangle meshes at $2048^{3}$ resolution. Twinkle3D advances high-fidelity geometry generation along four dimensions. First, while O-Voxel/FaithC offers high representational precision, it often suffers from poor surface quality and non-watertight geometry. We address both issues while retaining its $2048^{3}$-level precision, substantially improving surface quality and ensuring watertight mesh extraction. Second, we scale diffusion generation to sequences of up to 300K geometric tokens through a redesigned DiT architecture and large-scale distributed training optimizations, reducing training time per step from approximately ten minutes to ten seconds. Third, subsequent refinement cannot fully compensate for errors introduced during initial generation; we therefore strengthen both global shape and local detail in the initial generation stage, and the resulting single-stage model surpasses prior two-stage pipelines with $512^{3}$ refinement. Finally, we introduce a fine-grained image--3D cross-modal interaction mechanism that strengthens correspondence between visual evidence and geometric tokens, improving the recovery of object-specific structures. We evaluate geometric fidelity using alignment metrics derived from silhouettes and normal fields. Hi3D 3.0 outperforms four commercial systems across all reported metrics, recovering 82.1\% of inscribed characters at 98.2\% precision, compared with 21.7\% recall for the strongest competitor.
\end{abstract}

\vfill

{\hypersetup{hidelinks}\setcounter{tocdepth}{2}\tableofcontents}

\section{Introduction}
\label{sec:intro}

Image-to-3D generation has become increasingly capable of producing objects
that closely resemble the one depicted in a single input image. Early approaches lifted 2D
diffusion priors through per-object
optimization~\citep{poole2023dreamfusion,wang2023prolificdreamer}; these were
followed by multi-view
diffusion~\citep{liu2023zero123,shi2024mvdream,liu2024syncdreamer,long2024wonder3d,li2024era3d},
feed-forward
reconstruction~\citep{hong2024lrm,li2023instant3d,tochilkin2024triposr,xu2024instantmesh,wang2024crm},
and native 3D latent diffusion on Transformer
backbones~\citep{peebles2023dit,zhao2023michelangelo,zhang2023shape2vecset,zhang2024clay,wu2024direct3d},
which recent systems have scaled to production
quality~\citep{xiang2025trellis,li2025triposg,zhao2025hunyuan3d2,lai2025lattice}.
A parallel line of work generates artist-style meshes directly, either
autoregressively over vertices and
faces~\citep{nash2020polygen,siddiqui2024meshgpt,chen2024meshxl,chen2025meshanything,chen2025meshanythingv2,tang2025edgerunner,weng2025bpt,hao2024meshtron}
or in parallel with flow-based
models~\citep{he2025meshcraft,li2026meshflow,sun2026meshflow,xu2026meshyt2}.
Geometric fidelity has advanced alongside generative capability:
Sparc3D~\citep{li2025sparc3d} (Hi3D 1.0) showed that an $8\times$ compressed
sparse latent need not sacrifice reconstruction accuracy, and TRELLIS
2.0~\citep{xiang2025trellis2} removed the signed-distance-field intermediate
whose interpolation rounds sharp creases at the sampling scale. Given one image,
current systems reliably return a complete and coherent object whose overall
shape and proportions match the input.

Beyond agreement in overall shape, the outstanding challenge is to reproduce
the depicted object itself, including the specific geometry that defines it. This geometry
is small in extent, and its correctness is largely discrete: a face whose
interocular distance is altered by a few percent still resembles the input but
depicts a different person, and an inscribed character missing a stroke reads
as a different word or not at all. Such features, including inscriptions,
brand marks, and repeated structures, barely affect the silhouette, and
current systems frequently distort or lose them. We refer to agreement with
the input object at every scale, from overall proportion to detail near the
resolution limit, as \emph{object-specific fidelity}. Silhouette-based
measures are largely insensitive to it: on our benchmark, five current
systems lie within four points of silhouette IoU but differ substantially in
recovered fine structure (Sec.~\ref{sec:eval:main}).

In this report we present Twinkle3D, the geometry model of Hi3D 3.0, the
third generation of our image-to-3D series after Hi3D 1.0
(Sparc3D~\citep{li2025sparc3d}) and Hi3D 2.1~\citep{hi3d21}; comparisons with
released systems in Sec.~\ref{sec:eval} are reported under the Hi3D 3.0 name.
Given a single image, Twinkle3D generates a watertight triangle mesh at
$2048^{3}$ resolution (Figure~\ref{fig:teaser}) and advances high-fidelity
geometry generation along four dimensions, each addressing a point in the
pipeline at which object-specific fidelity is lost.

\paragraph{Representation and autoencoder.}
The representation and its autoencoder bound the detail available to the
generator. A representation that encodes a single surface patch per cell
cannot describe thin walls, closely spaced parts, or junctions at which
several surface sheets meet, and repairing a mesh that is not closed
introduces surface absent from the object.
O-Voxel/FaithC~\citep{xiang2025trellis2,luo2026faithc} admits several surface
sheets per cell and thereby offers high representational precision; in
practice, however, it often suffers from poor surface quality and
non-watertight geometry. We address both issues while retaining its
$2048^{3}$-level precision, substantially improving surface quality and
ensuring watertight mesh extraction without a repair step.
Sec.~\ref{sec:eval:vae} isolates this component through autoencoder
reconstruction.

\paragraph{Scaling generation to $2048^{3}$.}
A sample at $2048^{3}$ contains up to 300K geometric tokens, a sequence
length at which the full attention of a standard diffusion Transformer
becomes prohibitive; recent work reduces this cost with sparse
attention~\citep{wu2025direct3ds2}. We redesign the DiT architecture to
operate at this sequence length and introduce large-scale distributed
training optimizations, reducing the time per training step at thousand-GPU
scale from approximately ten minutes to ten seconds. This roughly $60\times$
reduction shortens a $10^{5}$-step schedule from close to two years to days.

\paragraph{Single-stage generation.}
High-resolution systems commonly adopt a coarse-to-fine factorization in
which a refinement stage adds detail conditioned on the output of a first
stage~\citep{xiang2025trellis,li2025sparc3d,xiang2025trellis2}. Refinement in
these pipelines adds detail relative to the first-stage surface and does not
revise its placement. First-stage errors consequently persist: a displacement
comparable to the feature size remains after refinement, and two neighbouring
surfaces whose displacement is comparable to their separation are merged.
Subsequent refinement cannot fully compensate for errors introduced during
initial generation; we therefore strengthen both global shape and local
detail in the initial generation stage. On its own, this single stage exceeds
the geometric accuracy of Sparc3D and TRELLIS 2.0 running a first stage
followed by $512^{3}$ refinement (Sec.~\ref{sec:eval:stage1}).

\paragraph{Image--3D cross-modal interaction.}
Object-specific features occupy a small area of the input image and are
therefore represented by few image tokens; recovering them requires the
generator to establish precise correspondence between this local visual
evidence and the geometric tokens that realize it. We introduce a
fine-grained image--3D cross-modal interaction mechanism that strengthens
this correspondence and improves the recovery of inscriptions, marks, and
repeated structures.

Evaluating object-specific fidelity presents a further difficulty. 3D
generation lacks an established automatic metric analogous to
FID~\citep{heusel2017fid} or CLIP score~\citep{hessel2021clipscore} in image
generation. The common substitute, scoring rendered images with a
vision-language model, conflates geometry with texture, material, and
lighting; when geometric differences are subtle, appearance dominates the
score. We therefore evaluate in the geometry domain
(Sec.~\ref{sec:eval:protocol}). Global alignment is measured by silhouette
IoU under a fitted camera pose, and detail alignment by a multi-scale
bandpass comparison of normal fields that penalizes spurious detail as
heavily as missing detail. Vision-language scoring is confined to multi-view
plausibility and shape aesthetics, and character-level fidelity is measured
on an OCR-scored inscription benchmark. Under this protocol, Hi3D 3.0
outperforms four commercial systems on all four metrics
(Table~\ref{tab:main}); the five systems span $0.04$ in silhouette IoU but
$0.11$ in detail alignment and $0.12$ in multi-view plausibility. On the
inscription benchmark, Hi3D 3.0 recovers $82.1\%$ of characters at $98.2\%$
precision, compared with $21.7\%$ recall and $62.0\%$ precision for the
strongest competitor (Sec.~\ref{sec:eval:text}). In a blind pairwise study,
it is rated equal to or better than each competitor in $78.0$--$89.8\%$ of
structure judgements and $88.1$--$95.5\%$ of geometric-detail judgements
(Sec.~\ref{sec:eval:user}).

In summary, this report identifies object-specific fidelity as a distinct
axis of image-to-3D quality that silhouette-based and appearance-based
evaluation does not capture, presents Twinkle3D, the geometry model of Hi3D
3.0, which is designed for it, and introduces a geometry-domain evaluation
protocol together with an OCR-scored inscription benchmark for measuring it.
We hope that this framing will shift attention from producing an object that
resembles the input to reproducing the object it depicts, and that the protocol will serve as a
baseline for subsequent work.

\section{Related Work}
\label{sec:related}

\paragraph{Native Mesh Generation.}
A line of work generates meshes directly as sequences of vertices and faces.
PolyGen~\citep{nash2020polygen} generates vertices and then faces with
autoregressive Transformers, and MeshGPT~\citep{siddiqui2024meshgpt}
generates sequences of learned triangle tokens with a decoder-only
Transformer. Subsequent work scales this formulation to larger models, data, and face
counts~\citep{chen2024meshxl,hao2024meshtron,xu2026meshmosaic}, conditions
generation on an input shape~\citep{chen2025meshanything,chen2025meshanythingv2},
and shortens or structures the token sequence through compressed, hierarchical,
and locality-aware
tokenizations~\citep{tang2025edgerunner,weng2025bpt,weng2025pivotmesh,lionar2025treemeshgpt,wang2025nautilus,song2026silksong,lin2026meshripple,wang2026face,xu2026meshweaver}.
Other work orders generation by level of
detail~\citep{zhang2025vertexregen,lei2025armesh}, accelerates decoding with
interleaved linear and full attention or speculative
decoding~\citep{wang2025iflame,shen2026flashmesh}, optimizes generation with
reinforcement learning~\citep{zhao2025deepmesh,zhou2026meshpro}, extends it to
quadrilateral meshes, part-wise generation, and native UV
charts~\citep{liu2026quadgpt,yang2026partdiffuser,xu2026strips}, and represents
meshes as plain text for a language model~\citep{wang2024llamamesh}.
Non-autoregressive methods generate all faces in parallel, with discrete
diffusion~\citep{alliegro2023polydiff}, continuous connectivity
representations~\citep{shen2024spacemesh}, or diffusion and flow
matching~\citep{he2025meshcraft,li2026meshflow,sun2026meshflow,wang2026nexus,xu2026meshyt2}.
These methods target compact, artist-like topology and controllable face
budgets. \Tw{} instead generates a dense surface at \res{2048} and extracts a
watertight mesh from it. It addresses whether the extracted surface
reproduces the input object; topology and face-budget control are outside the
scope of this report.

\paragraph{3D native shape generation.}
Early image-to-3D methods lift 2D priors, through per-object optimization with
a 2D diffusion model~\citep{poole2023dreamfusion,wang2023prolificdreamer},
consistent multi-view
synthesis~\citep{liu2023zero123,shi2024mvdream,liu2024syncdreamer,long2024wonder3d,li2024era3d},
or feed-forward
reconstruction~\citep{hong2024lrm,li2023instant3d,tochilkin2024triposr,xu2024instantmesh,wang2024crm}.
Native methods instead learn a generative model directly over 3D shapes.
Early native models generate point clouds or implicit
functions~\citep{zeng2022lion,nichol2022pointe,jun2023shape}.
3DShape2VecSet~\citep{zhang2023shape2vecset} encodes a shape as a set of
latent vectors decoded to a neural field, and diffusion over such latent sets
underlies Michelangelo~\citep{zhao2023michelangelo},
CLAY~\citep{zhang2024clay}, CraftsMan3D~\citep{li2024craftsman3d},
TripoSG~\citep{li2025triposg}, Step1X-3D~\citep{li2025step1x3d}, and
Hunyuan3D~2.0 and~2.1~\citep{zhao2025hunyuan3d2,hunyuan3d2025hunyuan3d21};
Dora~\citep{chen2025dora} improves the reconstruction of their autoencoders at
sharp edges, and MSVS-VAE~\citep{hao2026msvsvae} anchors the latent set at
multiple scales. Other latent forms include triplanes~\citep{chan2022eg3d,wu2024direct3d},
wavelets~\citep{hui2024makeashape,sanghi2024wala}, local signed-distance
grids~\citep{yariv2024mosaicsdf}, and primitives~\citep{chen2025topiaxl}.
Sparse voxel structures raise the attainable resolution.
XCube~\citep{ren2024xcube} generates sparse voxel hierarchies;
TRELLIS~\citep{xiang2025trellis} introduces structured latents with a
coarse-to-fine factorization that later systems
extend~\citep{lai2025lattice,lai2025hunyuan3d25,wang2026assetgen};
Direct3D-S2~\citep{wu2025direct3ds2} and SparseFlex~\citep{he2025sparseflex}
reach higher resolutions with sparse attention and sparse isosurface
representations; Ultra3D~\citep{chen2025ultra3d} refines sparse-voxel latents
with part attention; and LoG3D~\citep{yang2026log3d} partitions the volume into
local blocks to model shapes at \res{2048}. Sparc3D~\citep{li2025sparc3d} (Hi3D~1.0) shows that an
$8\times$ compressed sparse latent can preserve reconstruction accuracy, and
TRELLIS~2.0~\citep{xiang2025trellis2} removes the signed-distance intermediate
to keep sharp features.
Recent technical reports scale
these designs to high-fidelity, simulation-ready
assets~\citep{feng2025seed3d,gu2026seed3d2,jia2025ultrashape}, and
SAM~3D~\citep{sam3dteam2025sam3d} recovers objects from real-world images.
Other work encodes geometry and appearance in a single
latent~\citep{wu2025unilat3d}, aligns the generated object with the input view
through pixel-aligned features or joint pose
modeling~\citep{li2026pixal3d,huang2026cupid}, or generates each part of an
object separately~\citep{yang2025omnipart,ding2026fullpart,he2026unipart}.
\Tw{} belongs to this line of 3D native generation. It targets two
properties that resolution alone does not provide, cells crossed by several
surface sheets and watertight extraction without repair, and it revisits how
responsibility is divided between the first stage and refinement
(Sec.~\ref{sec:eval:stage1}).

\section{Experiments}
\label{sec:eval}

\subsection{Setup}
\label{sec:eval:protocol}

\paragraph{Benchmark.}
We evaluate on $33$ single-image inputs drawn from production requests. We
compare five systems: \Hi{}, Hi3D~2.1~\citep{hi3d21},
Meshy~7~\citep{meshy2026}, Tripo~3.1~\citep{tripo2026}, and Rodin
Gen~2.5~\citep{rodin2026}. Each system runs at its highest available
geometry resolution.

\paragraph{Metrics.}
We use four metrics, described next. We compute the first two from binary masks and
normal fields, so no texture, material, or lighting choice can affect
them. The last two use a vision-language judge, and only for questions
such judges answer reliably.

\paragraph{Global alignment (silhouette IoU).}
All systems share one set of
perspective intrinsics with a vertical field of view of $35^{\circ}$. For
each system we fit the camera pose by maximizing the intersection-over-union
between the rasterized mesh silhouette and the foreground mask of the
input. The score is the IoU at that pose. Fitting each system separately
avoids penalizing a convention difference in scale or orientation. This
metric measures projected occupancy only.

\paragraph{Detail alignment (normal bandpass).}
The reference is a monocular
normal estimate of the input image. The prediction is a normal rendering of
the mesh at the fitted pose. Both are restricted to the shared foreground
and compared through a multi-scale bandpass decomposition in normal space,
along two axes. ``Richness'' compares the energy in each band and
penalizes departures in both directions: over-smoothing and injected
detail are scored alike, so roughening a surface raises band energy without
raising the score. ``Envelope'' compares where each band's magnitude is
concentrated rather than its signed values, which tolerates the sub-pixel
registration error left by the pose fit. Because this metric depends on the
pose fit, we compute it only on inputs whose fit is high-confidence. We
note that this metric compares against a monocular normal estimate rather
than ground truth, so it measures agreement with an estimator and inherits
its biases. It ranks systems against a common reference and is not an
absolute measure of geometric error. We also note that systems whose poses
fit less reliably are partly excluded from it, and it is the metric that
most separates the field (Sec.~\ref{sec:eval:main}).

\paragraph{Multi-view plausibility (completeness).}
The two metrics above constrain only the
visible side. A vision-language model scores Blender back-view renderings
for whether the unseen side was completed, checking for holes, blank
faces, and undifferentiated surface, and returns a value in $[0,1]$. This is
a presence-or-absence judgement rather than a graded preference.

\paragraph{Shape aesthetics (Q-Judger).}
We render front and back views of each result with
a template built by professional artists, giving $328$ images. We score
them with Qwen-Image-Bench against a fixed three-level rubric: Fail
$\rightarrow 0$, Pass $\rightarrow 60$, Excel $\rightarrow 100$. This metric
is appearance-dependent by design.

\FloatBarrier
\subsection{Main Results}
\label{sec:eval:main}

\begin{table}[h]
  \caption{\textbf{Main results on the $33$-input benchmark.} \Hi{} has
  the best value on all four metrics. Global alignment is silhouette IoU.
  Detail alignment is the appearance-independent normal bandpass score,
  computed on the high-confidence-pose subset. Multi-view plausibility is
  the vision-language completeness score in $[0,1]$. Shape aesthetics (Q-Judger) is
  the Qwen-Image-Bench score in $[0,100]$. All systems run at their highest
  geometry resolution. Best per column in \textbf{bold}.}
  \label{tab:main}
  \centering
  \small
  \begin{tabular}{lcccc}
    \toprule
    Model & \begin{tabular}[c]{@{}c@{}}Global\\Alignment $\uparrow$\end{tabular}
          & \begin{tabular}[c]{@{}c@{}}Detail\\Alignment $\uparrow$\end{tabular}
          & \begin{tabular}[c]{@{}c@{}}Multi-view\\Plausibility $\uparrow$\end{tabular}
          & \begin{tabular}[c]{@{}c@{}}Shape Aesthetics\\(Q-Judger) $\uparrow$\end{tabular} \\
    \midrule
    Rodin Gen 2.5 & 0.8787 & 0.2645 & 0.663 & 46.67 \\
    Tripo 3.1     & 0.8953 & 0.2826 & 0.722 & 50.16 \\
    Meshy 7       & 0.9013 & 0.3694 & 0.638 & 50.00 \\
    Hi3D 2.1      & 0.9038 & 0.2813 & 0.636 & 50.30 \\
    \midrule
    \textbf{Hi3D 3.0} & \textbf{0.9184} & \textbf{0.3782} & \textbf{0.757} & \textbf{50.91} \\
    \bottomrule
  \end{tabular}
\end{table}

Table~\ref{tab:main} compares the five systems. All five run at their
highest geometry resolution and are scored by the same protocol, so the
differences can only be attributed to the generated geometry. \Hi{} obtains
the best value on all four metrics. We have three observations from
Table~\ref{tab:main}.

First, silhouettes are saturated but detail is not. Global alignment spans
$0.879$ to $0.918$ across the five systems ($4\%$ of the column mean), but
detail alignment spans $0.265$ to $0.378$ ($36\%$ of its mean). The field
also splits on detail alignment rather than spreading evenly: \Hi{}
($0.3782$) and Meshy~7 ($0.3694$) sit together, and Tripo~3.1 ($0.2826$),
Hi3D~2.1 ($0.2813$), and Rodin Gen~2.5 ($0.2645$) sit $0.087$ to $0.114$
below them. Because the metric penalizes excess band energy as heavily as
deficit, this gap reflects fine structure that agrees with the input, not
merely fine structure that is present.

Second, completeness is a separate axis. Multi-view plausibility varies
independently of the two alignment metrics. \Hi{} and Meshy~7, the top two
systems on detail alignment, differ by $0.119$ on completeness.
Hi3D~2.1 and Meshy~7 differ by $0.088$ on detail alignment but by only
$0.002$ on completeness. We also note that completeness is far from its ceiling for
every system, including ours at $0.757$.

Last, aesthetics do not separate the field. Four of the five systems fall
within one point of each other on shape aesthetics. We attribute this to the three-level rubric, which
saturates once outputs reliably reach the middle level. We report this
metric to cover the appearance dimension. The widest spreads come from
detail alignment and completeness, the two metrics that measure geometry
beyond projected occupancy.

\FloatBarrier
\subsection{Inscription Fidelity}
\label{sec:eval:text}

Inscriptions give a discrete measure of fidelity. A character either reads
back or it does not, and the judgement does not depend on rendering style.
A model cannot score well by producing text-like relief.

\paragraph{Protocol.} The test set covers Chinese and English inscriptions,
engravings, and brand marks. We render each system's output under clay
shading and read it with PaddleOCR~\citep{du2020ppocr} against a shared
scene-level ground truth of $526$ characters, aligned character by
character. We report recall (ground-truth characters recovered), precision
(the share of everything OCR reads that was present), and their harmonic mean
F1. We also report the miss rate (ground-truth characters OCR cannot locate, i.e.,
geometry below legibility) and the hallucination rate (invented characters
as a share of ground truth plus extras).

\begin{table}[h]
  \caption{\textbf{Inscription fidelity.} \Hi{} recovers $82.1\%$ of the
  $526$ ground-truth characters at $98.2\%$ precision; the best peer system
  recovers $21.7\%$. Measured with PaddleOCR on clay-shaded renderings
  against a shared ground truth. Recall is the share of ground-truth characters recovered. Precision is
  $\mathrm{correct}/(\mathrm{correct}+\mathrm{wrong}+\mathrm{extra})$. F1 is
  their harmonic mean. Miss counts ground-truth characters OCR cannot
  locate, and Halluc.\ counts invented characters as a share of ground
  truth plus extras. Precision and F1 are derived from the counts in the
  last column. Peer systems run at their latest version and highest
  geometry resolution. Best per column in \textbf{bold}.}
  \label{tab:text}
  \centering
  \small
  \begin{tabular}{lcccccc}
    \toprule
    Model & Recall $\uparrow$ & Precision $\uparrow$ & F1 $\uparrow$
          & Miss $\downarrow$ & Halluc.\ $\downarrow$
          & \begin{tabular}[c]{@{}c@{}}Missed / Wrong /\\Extra\end{tabular} \\
    \midrule
    Hi3D 2.1  &  0.0\% &  0.0\% &  0.0\% & 99.2\% & 4.5\%  & 522 / 4 / 25 \\
    Rodin Gen 2.5 &  4.6\% & 21.8\% &  7.5\% & 88.0\% & 8.2\%  & 463 / 39 / 47 \\
    Tripo 3.1 &  7.4\% &  9.8\% &  8.4\% & 71.3\% & 32.0\% & 375 / 112 / 247 \\
    Meshy 7   & 21.7\% & 62.0\% & 32.1\% & 76.0\% & 9.9\%  & 400 / 12 / 58 \\
    \midrule
    \textbf{Hi3D 3.0} & \textbf{82.1\%} & \textbf{98.2\%} & \textbf{89.4\%}
                      & \textbf{16.5\%} & \textbf{0.2\%}  & 87 / 7 / 1 \\
    \bottomrule
  \end{tabular}
\end{table}

\begin{figure}[t]
  \centering
  \setlength{\rowh}{84pt}\setlength{\tabcolsep}{2pt}%
  \renewcommand{\arraystretch}{1}%
  \begin{tabular}{@{}cc@{\hspace{10pt}}cc@{}}
    \setlength{\cellw}{1.5401\rowh}\colhdr{\itshape Input} &
    \setlength{\cellw}{1.5401\rowh}\colhdrours{Hi3D 3.0} &
    \setlength{\cellw}{1.0716\rowh}\colhdr{\itshape Input} &
    \setlength{\cellw}{1.0716\rowh}\colhdrours{Hi3D 3.0} \\[2pt]
    \setlength{\cellw}{1.5401\rowh}\colimg{\includegraphics[width=\cellw]{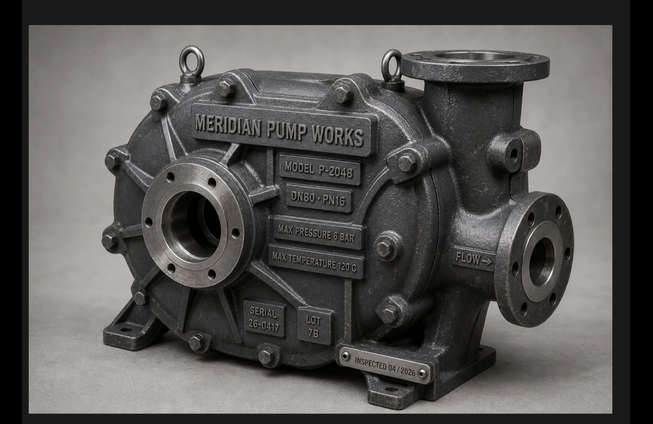}} &
    \setlength{\cellw}{1.5401\rowh}\colimgours{\includegraphics[width=\cellw]{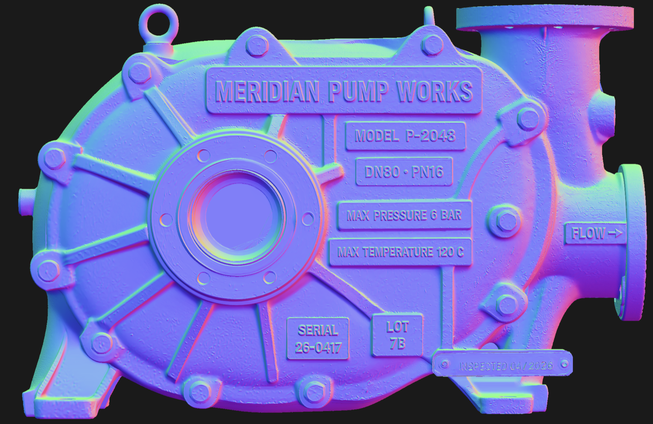}} &
    \setlength{\cellw}{1.0716\rowh}\colimg{\includegraphics[width=\cellw]{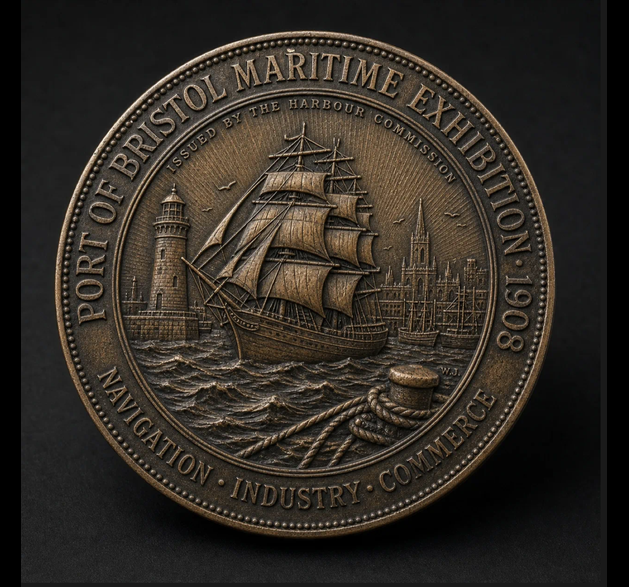}} &
    \setlength{\cellw}{1.0716\rowh}\colimgours{\includegraphics[width=\cellw]{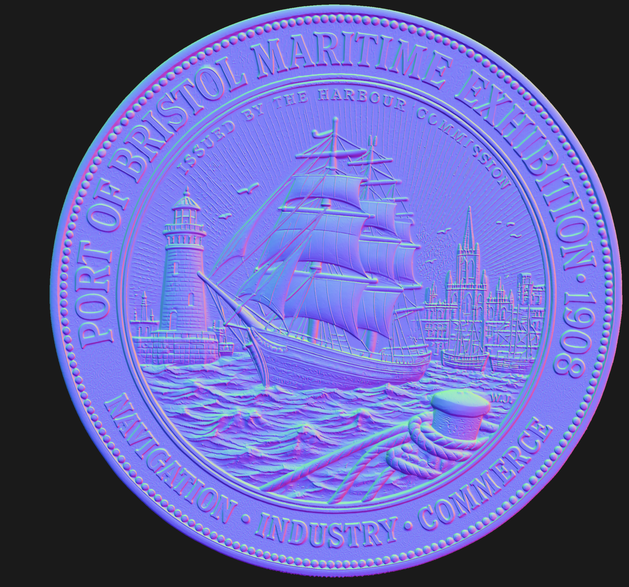}} \\
  \end{tabular}
  \caption{\textbf{Text-bearing objects.} The specification plates and the
  rim lettering are reproduced as geometry. Each pair shows the input
  photograph and the \Tw{} output under normal shading, so nothing in the
  comparison depends on texture or material. Left: an industrial pump with a
  model number, pressure and temperature ratings, serial and lot plates, and
  a cast flow arrow, alongside machined bolt bosses, the flange bore, and the
  lifting eye. Right: a commemorative medal with lettering curved along its
  rim over a low-relief scene.}
  \label{fig:twin-text}
\end{figure}

\begin{figure}[t]
  \centering
  \includegraphics[width=\linewidth]{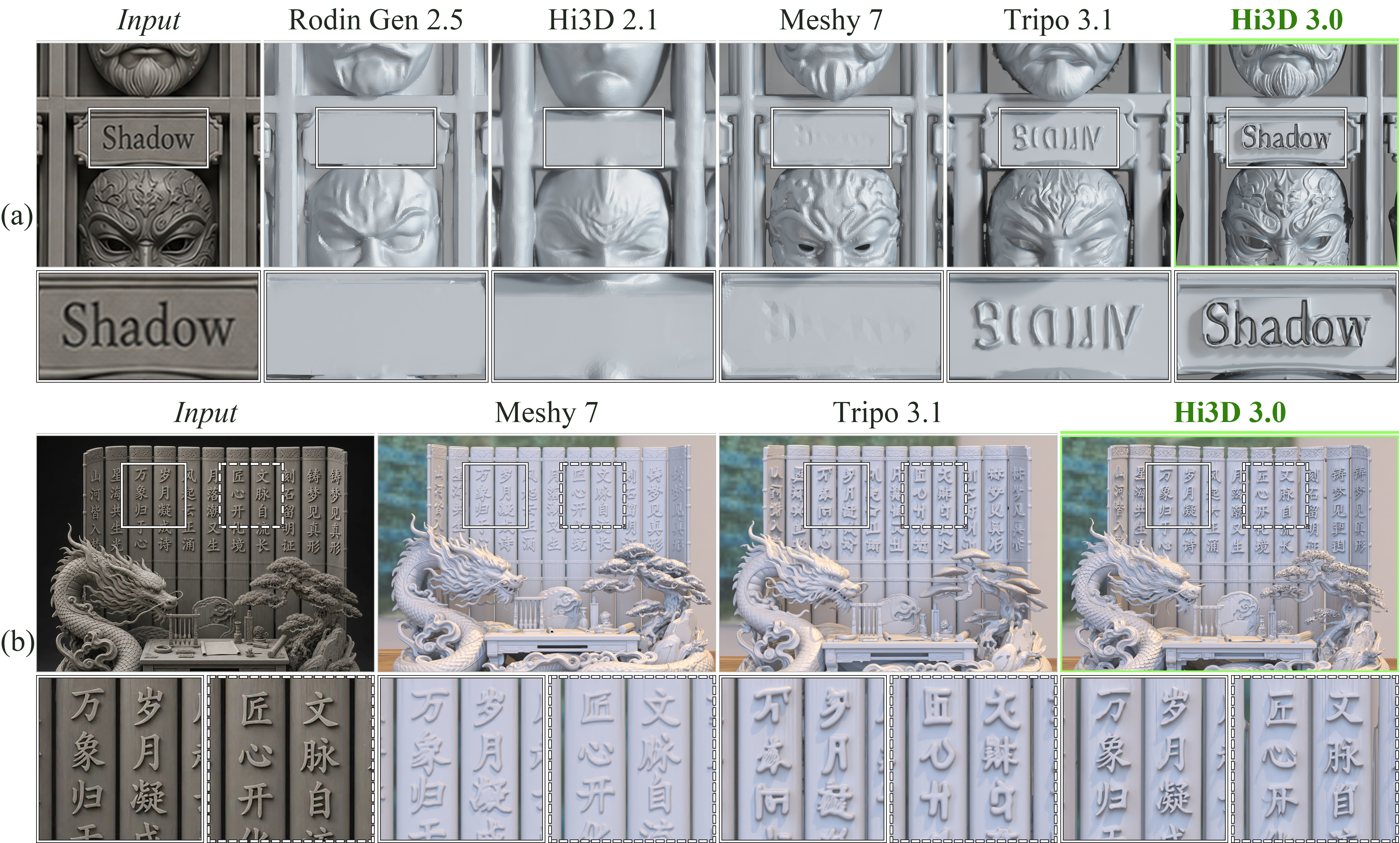}
  \caption{\textbf{Inscription recovery.} The white boxes in each view are enlarged beneath it. (a)~The engraved name plate ``Shadow'' on the mask cabinet, from one frame of a close-up camera path, the same frame for every system, under clay shading. Only \Hi{} reproduces the word. Tripo~3.1 forms garbled letter-like strokes, the hallucination mode counted in Table~\ref{tab:text}; Meshy~7 leaves faint marks; Rodin Gen~2.5 returns a blank plate; and Hi3D~2.1 forms only a plain bar. (b)~Carved calligraphy on a scholar's diorama, from the same turntable frame for each system. Rodin Gen~2.5 and Hi3D~2.1 were not rendered for this input. The solid and dashed boxes cover two pairs of columns, enlarged left and right. \Hi{} forms every enlarged character with sharp, deeply cut strokes. Meshy~7 forms the same characters with shallower and softer strokes, and Tripo~3.1 forms pseudo-characters that do not read as the input. Outside the boxes, two characters in the \Hi{} result are malformed.}
  \label{fig:text-compare}
\end{figure}

\paragraph{Results.} Table~\ref{tab:text} shows the results. \Hi{} reads
back $82.1\%$ of ground-truth characters. The best peer system (Meshy~7)
reaches $21.7\%$, and three of the four remain below $8\%$. Meshy~7 is also
the closest competitor on detail alignment in Table~\ref{tab:main}. Hi3D~2.1 recovers
none: $99.2\%$ of its characters fall below legibility. This suggests that the
inscription scale lies beneath that system's effective resolution.

Precision separates the systems further and distinguishes two failure
modes. Of everything PaddleOCR reads from a \Hi{} result, $98.2\%$ is a
character that is present, vs.\ $62.0\%$ for Meshy~7 and $9.8\%$ for
Tripo~3.1. A model that omits illegible text keeps precision high while
losing recall. A model that invents text-like relief loses both. So Tripo~3.1 and Rodin Gen~2.5 land within a point of each other on F1 ($8.4$
vs.\ $7.5$), although Tripo~3.1 recovers about $60\%$ more characters.
Across the whole benchmark, \Hi{} produces one invented character.

Figure~\ref{fig:twin-text} shows the same property outside the OCR setting,
on two objects whose identity is carried by cast and engraved detail. These are the
specification plates of an industrial pump (Figure~\ref{fig:twin-text}, left) and
the rim lettering of a medal (Figure~\ref{fig:twin-text}, right). Both are
normal-shaded. Figure~\ref{fig:text-compare} compares systems on two
inscription-bearing inputs. In Figure~\ref{fig:text-compare}(a),
three systems return the plate without legible content, and Tripo~3.1
fills it with letter-like strokes that do not read as the word. This is the
behaviour the hallucination column measures. In
Figure~\ref{fig:text-compare}(b), Meshy~7 recovers the carved characters with
shallower strokes, consistent with its nontrivial character rate, and
Tripo~3.1 again forms text-like relief that does not read as the input.

\subsection{Human Evaluation}
\label{sec:eval:user}

We ran a blind pairwise study comparing \Hi{} against each of the four
peer systems. Raters judged two dimensions separately (overall structure
and geometric detail quality) and returned one of three verdicts per
comparison: \Hi{} better, the same, or \Hi{} inferior.

\begin{table}[h]
  \caption{\textbf{Blind pairwise human study.} \Hi{} is rated the same or
  better in $78.0$--$89.8\%$ of structure judgements and $88.1$--$95.5\%$ of
  detail judgements. Each entry is the share of
  judgements in which \Hi{} was rated the same as, or better than, the
  comparison system.}
  \label{tab:userstudy}
  \centering
  \small
  \begin{tabular}{lcc}
    \toprule
    \Hi{} vs. & Structure (same or better) & Geometric Detail (same or better) \\
    \midrule
    Hi3D 2.1  & 89.0\% & 94.7\% \\
    Meshy 7   & 78.0\% & 91.1\% \\
    Rodin Gen 2.5 & 89.8\% & 95.5\% \\
    Tripo 3.1 & 79.5\% & 88.1\% \\
    \bottomrule
  \end{tabular}
\end{table}

Table~\ref{tab:userstudy} reports the same-or-better share. \Hi{} is rated the same
or better in $78.0$--$89.8\%$ of structure judgements and $88.1$--$95.5\%$
of detail judgements. Against every system the detail margin exceeds the
structure margin, from $5.7$ points (Rodin Gen~2.5 and Hi3D~2.1) to $13.1$ points
(Meshy~7). This agrees with Table~\ref{tab:main} on two points: (i)~the difference
between systems lies in detail rather than in structure, and (ii)~Meshy~7
is the closest competitor, on detail alignment there and on structure
preference here.

\FloatBarrier
\subsection{Autoencoder Reconstruction}
\label{sec:eval:vae}

\begin{figure}[t]
  \centering
  \gridsetup{0.30\linewidth}
  \begin{tabular}{@{}ccc@{}}
    \colhdr{Sparc3D} & \colhdr{TRELLIS 2.0} & \colhdrours{Twinkle3D} \\
  \end{tabular}\\[2pt]
  \begin{subfigure}{\linewidth}
    \centering
    \begin{tabular}{@{}ccc@{}}
      \colimg{\includegraphics[width=\cellw]{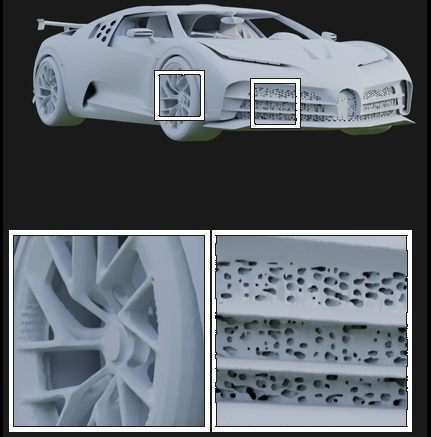}} &
      \colimg{\includegraphics[width=\cellw]{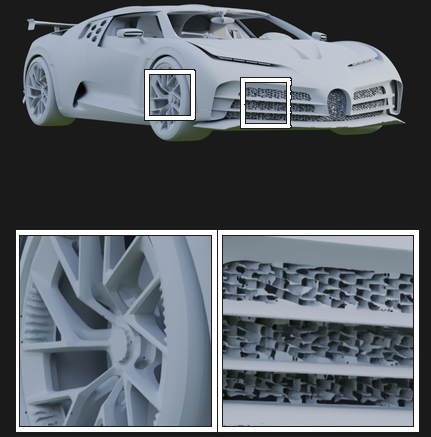}} &
      \colimgours{\includegraphics[width=\cellw]{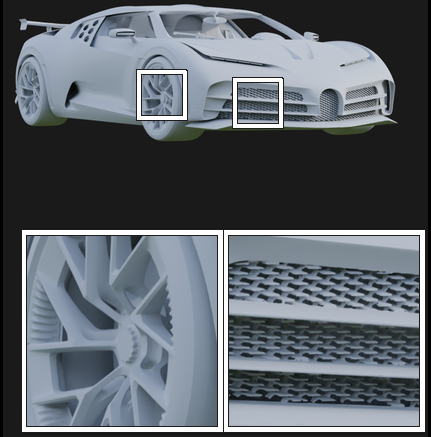}} \\
    \end{tabular}
    \caption{Wheel spokes and front-grille lattice.}
    \label{fig:vae-car}
  \end{subfigure}
  \\[0.4em]
  \begin{subfigure}{\linewidth}
    \centering
    \begin{tabular}{@{}ccc@{}}
      \colimg{\includegraphics[width=\cellw]{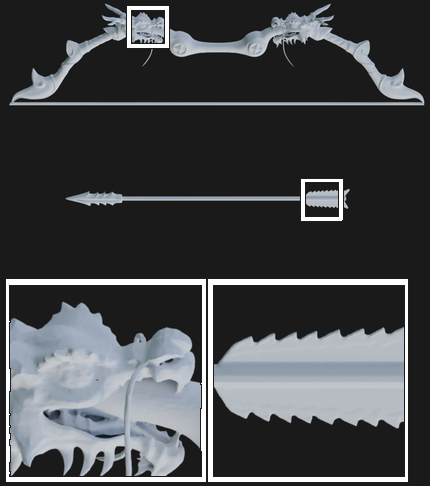}} &
      \colimg{\includegraphics[width=\cellw]{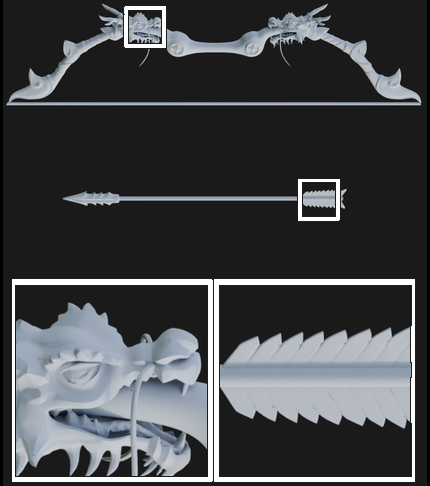}} &
      \colimgours{\includegraphics[width=\cellw]{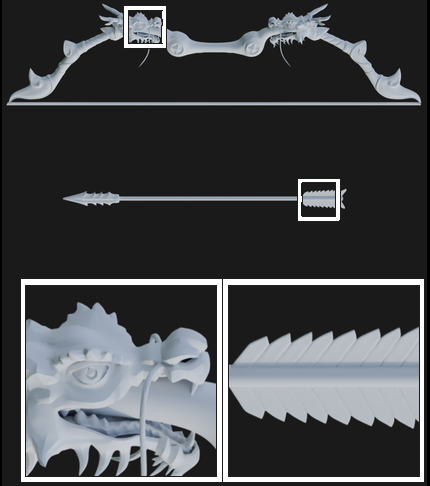}} \\
    \end{tabular}
    \caption{Skeletal bow and barbed arrow shaft.}
    \label{fig:vae-bow}
  \end{subfigure}
  \caption{\textbf{Autoencoder reconstruction.} We pass the same input mesh
  through the autoencoders of Sparc3D~\citep{li2025sparc3d},
  TRELLIS~2.0~\citep{xiang2025trellis2}, and \Tw{}. Only \Tw{} keeps the
  grille periodic and in phase, and Sparc3D alone tapers the arrow barbs into
  a cone. Columns are as labelled at
  the top throughout. White boxes mark the regions enlarged
  beneath each column. Both cases require cells that carry more than one
  surface sheet (Sec.~\ref{sec:intro}).}
  \label{fig:vae-recon}
\end{figure}

This experiment isolates the representation and its autoencoder
(Sec.~\ref{sec:intro}). Figure~\ref{fig:vae-recon} compares
reconstruction of the same input mesh through
Sparc3D~\citep{li2025sparc3d}, TRELLIS~2.0~\citep{xiang2025trellis2}, and
\Tw{} on two objects whose identity is carried by fine structure.

The front grille in Figure~\ref{fig:vae-car} is a periodic array of thin,
closely spaced elements: the multi-sheet-per-cell case. Sparc3D returns it
as irregular perforation and loses the periodicity. TRELLIS~2.0 recovers
the bar structure but not its spacing or thickness. \Tw{} keeps the array
periodic and in phase across the full width of the grille. All three
outputs carry fine-scale content, but only one carries fine-scale content
that matches the reference. This is the distinction that the richness term
of the detail metric registers. Figure~\ref{fig:vae-bow} shows a
single thin element: the arrow's barbs are a repeated profile along one
shaft. Sparc3D tapers them into a cone, and TRELLIS~2.0 and \Tw{} keep them
separate.

\subsection{One Stage vs.\ Two Stages}
\label{sec:eval:stage1}

\begin{figure}[p]
  \centering
  \gridsetup{0.28\linewidth}
  \begin{tabular}{@{}ccc@{}}
    \colhdr{Sparc3D\hdrsub{stage 1 + stage 2 (\res{512})}} &
    \colhdr{TRELLIS 2.0\hdrsub{stage 1 + stage 2 (\res{512})}} &
    \colhdrours{Twinkle3D\hdrsub{stage 1 only}} \\
  \end{tabular}\\[3pt]
  \begin{subfigure}{\linewidth}
    \centering
    \begin{tabular}{@{}ccc@{}}
      \colimg{\includegraphics[width=\cellw]{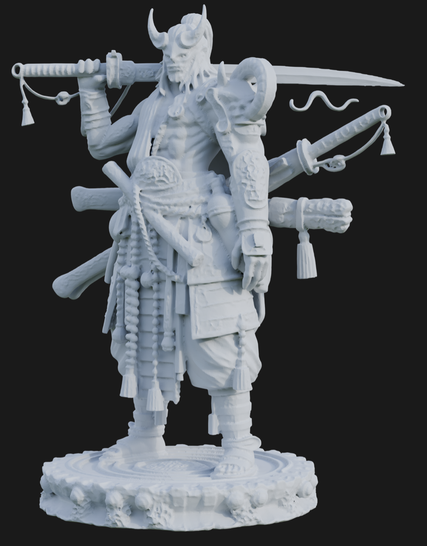}} &
      \colimg{\includegraphics[width=\cellw]{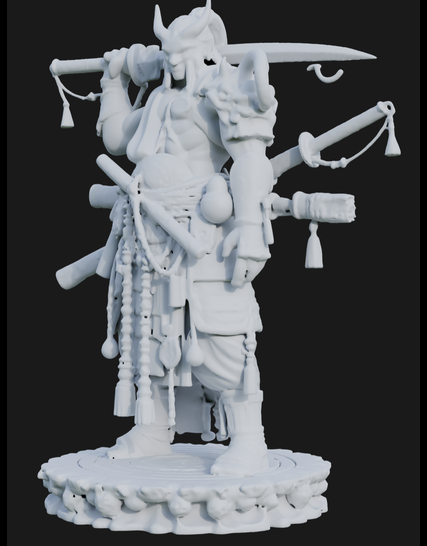}} &
      \colimgours{\includegraphics[width=\cellw]{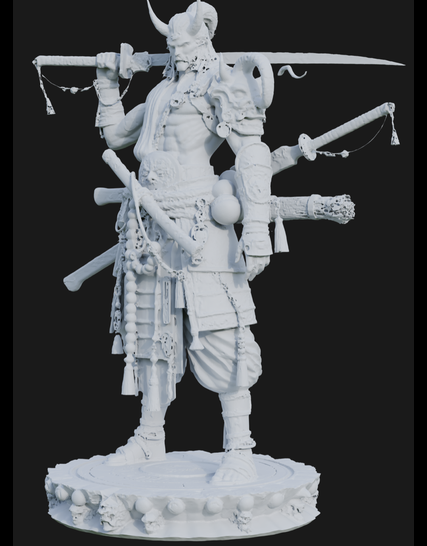}} \\
    \end{tabular}
    \caption{Armoured figure: cords and tassels.}
    \label{fig:stage1-warrior}
  \end{subfigure}
  \\[0.35em]
  \begin{subfigure}{\linewidth}
    \centering
    \begin{tabular}{@{}ccc@{}}
      \colimg{\includegraphics[width=\cellw]{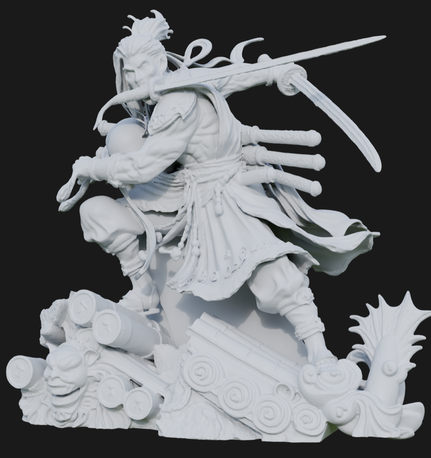}} &
      \colimg{\includegraphics[width=\cellw]{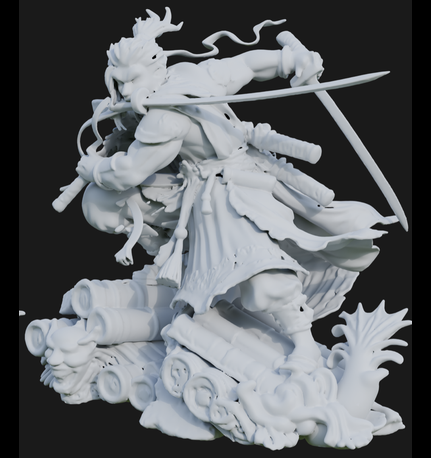}} &
      \colimgours{\includegraphics[width=\cellw]{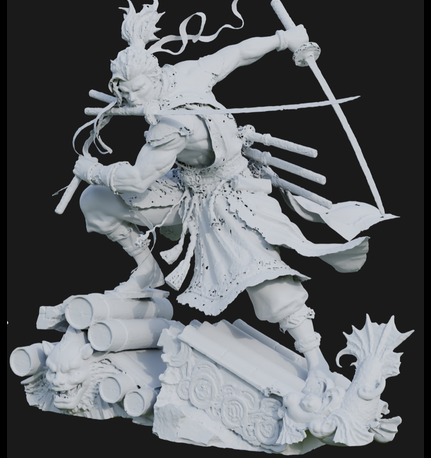}} \\
    \end{tabular}
    \caption{Swordsman: perforated robe and separate blades.}
    \label{fig:stage1-swordsman}
  \end{subfigure}
  \\[0.35em]
  \begin{subfigure}{\linewidth}
    \centering
    \begin{tabular}{@{}ccc@{}}
      \colimg{\includegraphics[width=\cellw]{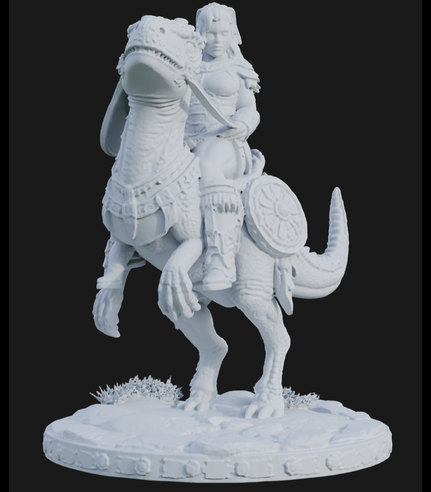}} &
      \colimg{\includegraphics[width=\cellw]{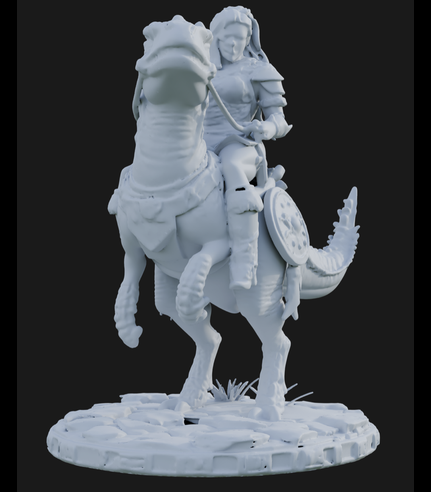}} &
      \colimgours{\includegraphics[width=\cellw]{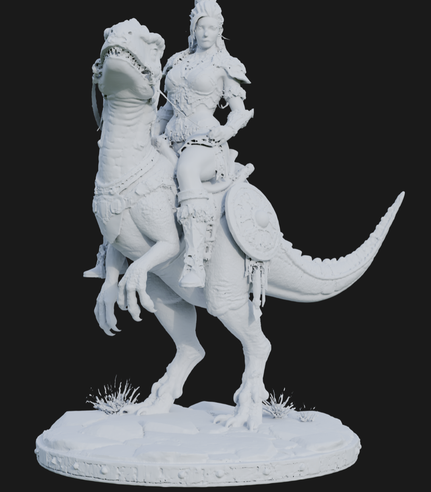}} \\
    \end{tabular}
    \caption{Mounted rider: saddle, reins and harness.}
    \label{fig:stage1-rider}
  \end{subfigure}
  \caption{\textbf{One stage vs.\ two.} We compare the standalone first stage of
  \Tw{} with a first stage plus \res{512} refinement in
  Sparc3D~\citep{li2025sparc3d} and TRELLIS~2.0~\citep{xiang2025trellis2},
  on three inputs. Columns are as labelled at the top throughout. \Tw{}
  runs one stage; both baselines run two. In every case the baselines fuse
  structures that should be separate, and the single-stage result keeps
  them distinct.}
  \label{fig:stage1-compare}
\end{figure}

This experiment tests the first-stage design of
Sec.~\ref{sec:intro}. We compare the standalone first stage of \Tw{}
against Sparc3D and TRELLIS~2.0 running a first stage plus \res{512}
refinement (Figure~\ref{fig:stage1-compare}). \Tw{} runs one stage; both
baselines run two.

On all three inputs, the single stage recovers structure that the
refined baselines do not. In
Figure~\ref{fig:stage1-swordsman}, the perforated pattern of the robe, the
separation between individual blades, and the layered scrollwork of the
base appear as distinct surfaces rather than fused ones. The same holds for
the cords and tassels in Figure~\ref{fig:stage1-warrior} and the saddle and
harness in Figure~\ref{fig:stage1-rider}. Fusion is the signature we expect
from Sec.~\ref{sec:intro}. When two nearby boundaries are each displaced
by an amount comparable to the gap between them, refinement adds surface
detail without correcting where the boundaries sit, and the gap closes.

\subsection{Qualitative Results}
\label{sec:eval:qualitative}

\begin{figure}[p]
  \centering
  \includegraphics[width=0.925\linewidth]{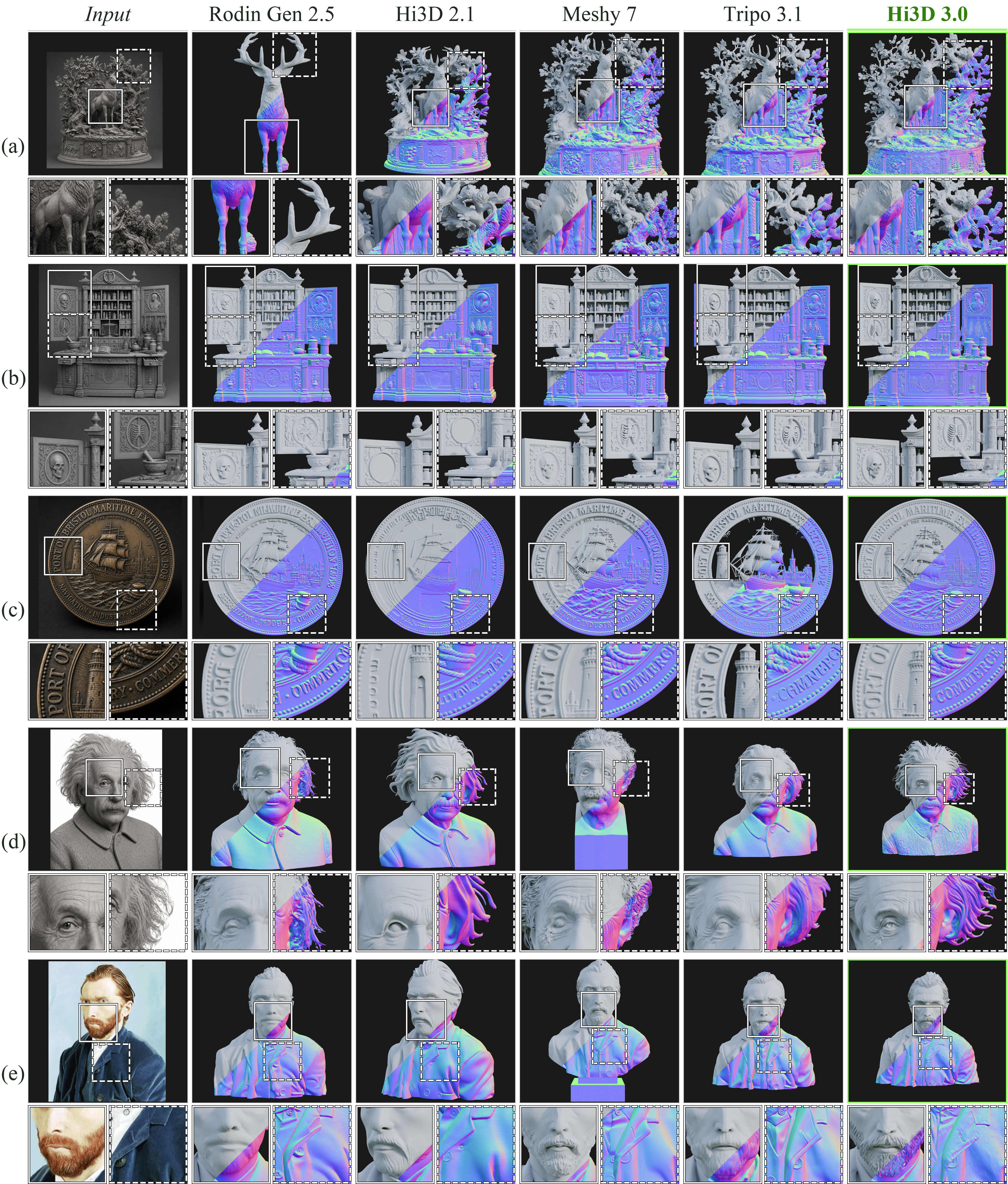}
  \caption{\textbf{Five-system comparison with two close-ups per input.}
  Each result is split diagonally, clay shading above and normal shading
  below. Beneath each view, the left close-up enlarges the solid white box
  and the right close-up the dashed white box; each box marks the same part
  of the object in every column, and all close-ups share one source
  resolution. (a)~Front legs; the right-hand tree. Rodin Gen~2.5 returns the stag alone, without the niche, foliage, or base, so its right close-up shows only an antler. \Hi{}, Meshy~7, and Tripo~3.1 form separate leaves and cones on the tree, which Hi3D~2.1 merges into lumpy clusters. (b)~Skull and ribcage medallions on the left door. Hi3D~2.1 leaves both medallions blank; \Hi{} carves the eye sockets and teeth of the skull and separate ribs along the spine. (c)~The lighthouse with the rim lettering ``PORT OF''; the rim lettering ``COMMERCE''. Meshy~7 and \Hi{} form both words and a windowed lighthouse. Tripo~3.1 misshapes ``COMMERCE'' and cuts away the inner field of the medal, leaving the lighthouse free-standing. Rodin Gen~2.5 forms ``PORT OF'' but garbles ``COMMERCE'' and forms no lighthouse. Hi3D~2.1 forms illegible glyphs and a plain tower. (d)~Eye and forehead; side hair. \Hi{} forms separate
  tapered strands and a carved iris, where the other systems form clumps,
  ribbons, or smooth sheets; Meshy~7 also truncates the bust at the neck.
  (e)~Beard; collar and buttons. \Hi{} and Meshy~7 form beard strands and
  curls, and Rodin Gen~2.5 forms no beard; all five systems reproduce the
  collar and buttons. Meshy~7 adds a plinth that is not in the input to
  both busts.}
  \label{fig:cmp-all}
\end{figure}

\begin{figure}[p]
  \centering
  \includegraphics[width=0.95\linewidth]{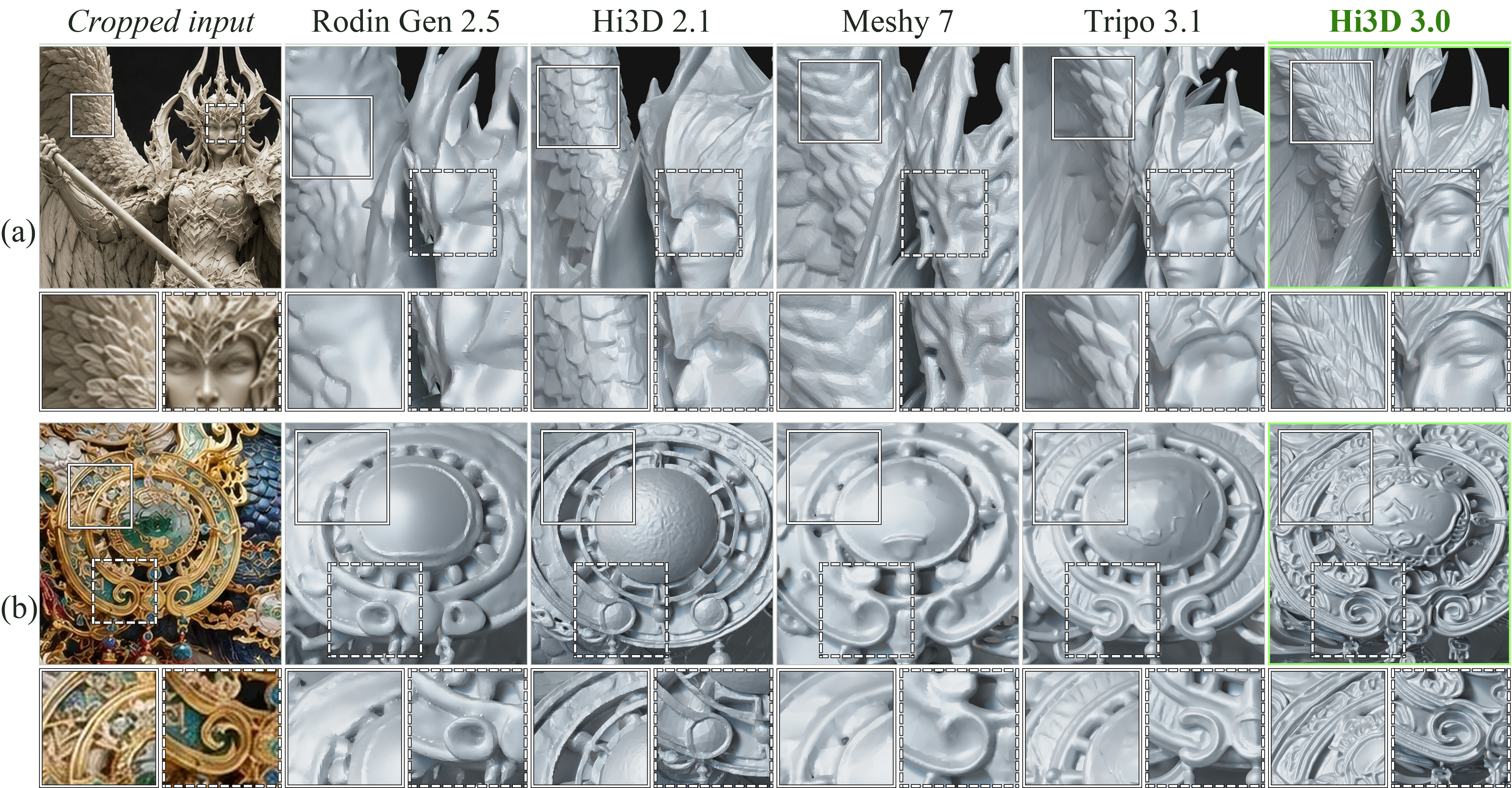}
  \caption{\textbf{Close-up comparison on ornament and relief.} Left: the input image. Right: one frame of the same close-up camera path for each system, the same frame for every system, under clay shading. The solid and dashed white boxes are enlarged beneath, left and right. (a)~Winged figure: the wing feathers and the face with its crown band. \Hi{} separates individual feathers and carves the eyelids and the edges of the crown band. Tripo~3.1 forms smooth feather lobes and a clean but soft face, Hi3D~2.1 faceted feather plates, Meshy~7 noisy ridges and a face fused with the crown, and Rodin Gen~2.5 a nearly smooth wing and a melted crown band. (b)~Qilin: the filigree band of a medallion and the C-scroll volute on its lower edge. \Hi{} fills the band with scroll relief and forms a spiral volute with nested curls. Tripo~3.1 forms radial grooves and a single tube-profile scroll, Meshy~7 an empty band and a soft curl, and Rodin Gen~2.5 and Hi3D~2.1 no scroll.}
  \label{fig:qual-compare-1}

  \vspace{14pt}
  \includegraphics[width=0.95\linewidth]{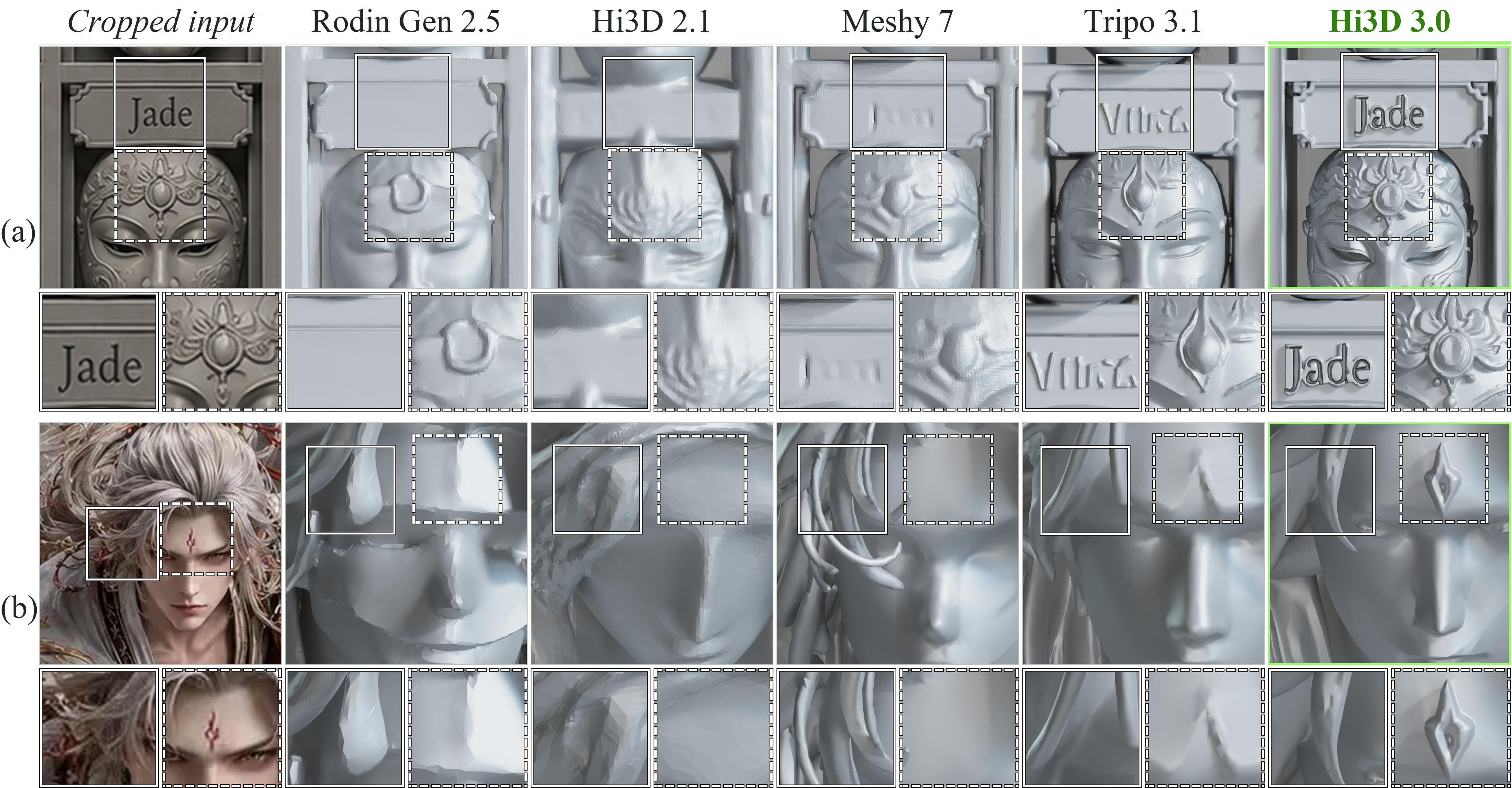}
  \caption{\textbf{Close-up comparison on inscriptions and identifying marks.} Conventions as in Figure~\ref{fig:qual-compare-1}. (a)~Mask cabinet: the ``Jade'' name plate and the forehead ornament of the mask beneath it. Only \Hi{} reproduces the engraved word, and it forms a crown of pointed petals around a jewel. Tripo~3.1 forms letter-like strokes that do not read as the word and a plain teardrop, Meshy~7 faint marks and a low bump, Rodin Gen~2.5 a blank plate and a single ring, and Hi3D~2.1 no plate and a ridge. (b)~Fox immortal: a lock of hair over the temple and the forehead mark, which appears in the input as a small red mark. \Hi{} forms notched locks and recovers the mark as a raised diamond with a recessed centre. Tripo~3.1 forms smooth strands and a shallow crease, Meshy~7 tube-shaped strands and no mark, and Rodin Gen~2.5 and Hi3D~2.1 faceted hair and no mark.}
  \label{fig:qual-compare-2}
\end{figure}

\begin{figure}[t]
  \centering
  \begin{subfigure}[t]{0.492\linewidth}
    \centering
    \includegraphics[width=\linewidth]{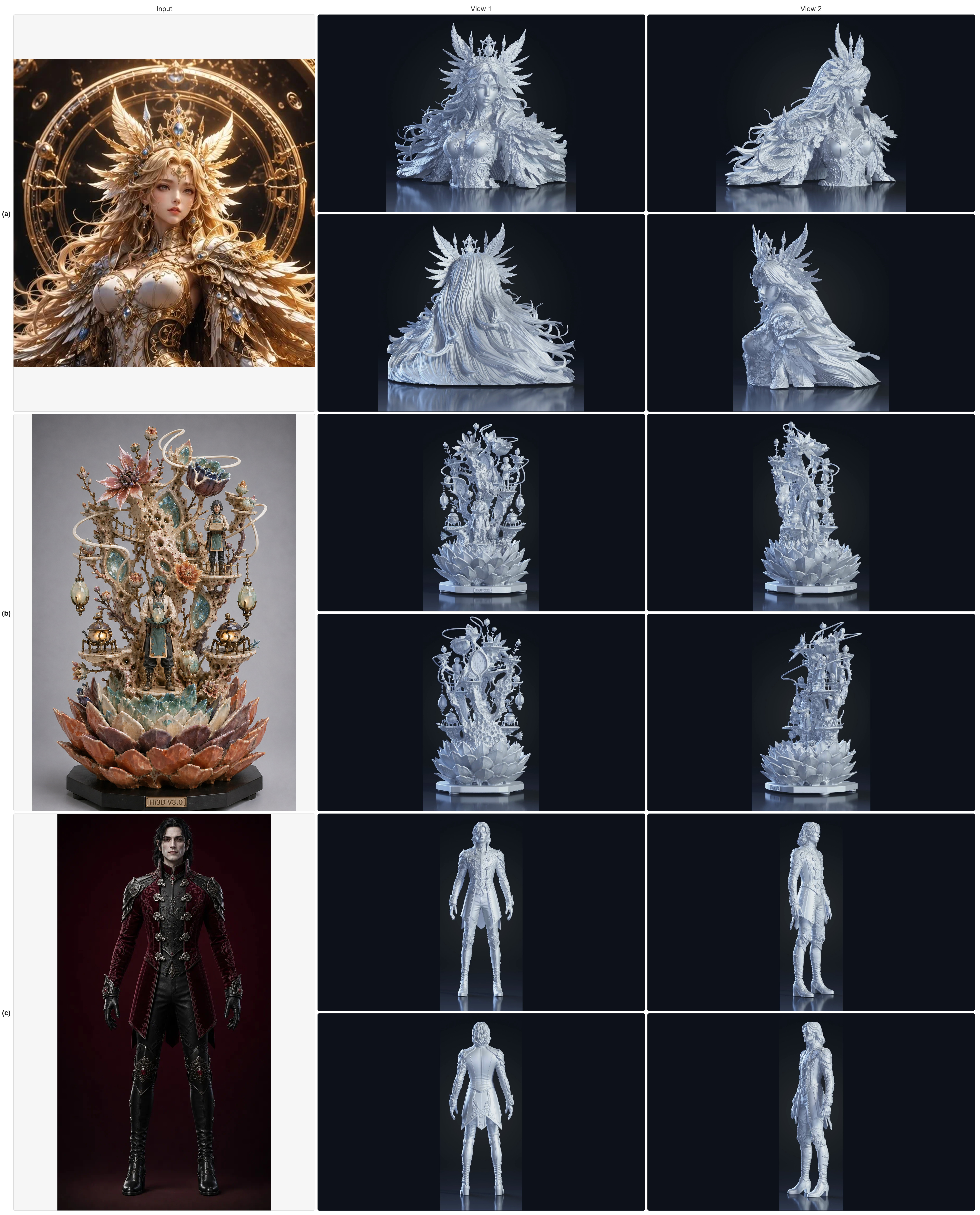}
    \caption{Ornate bust; multi-object diorama; full-body character.}
    \label{fig:detailed-show-a}
  \end{subfigure}\hfill
  \begin{subfigure}[t]{0.492\linewidth}
    \centering
    \includegraphics[width=\linewidth]{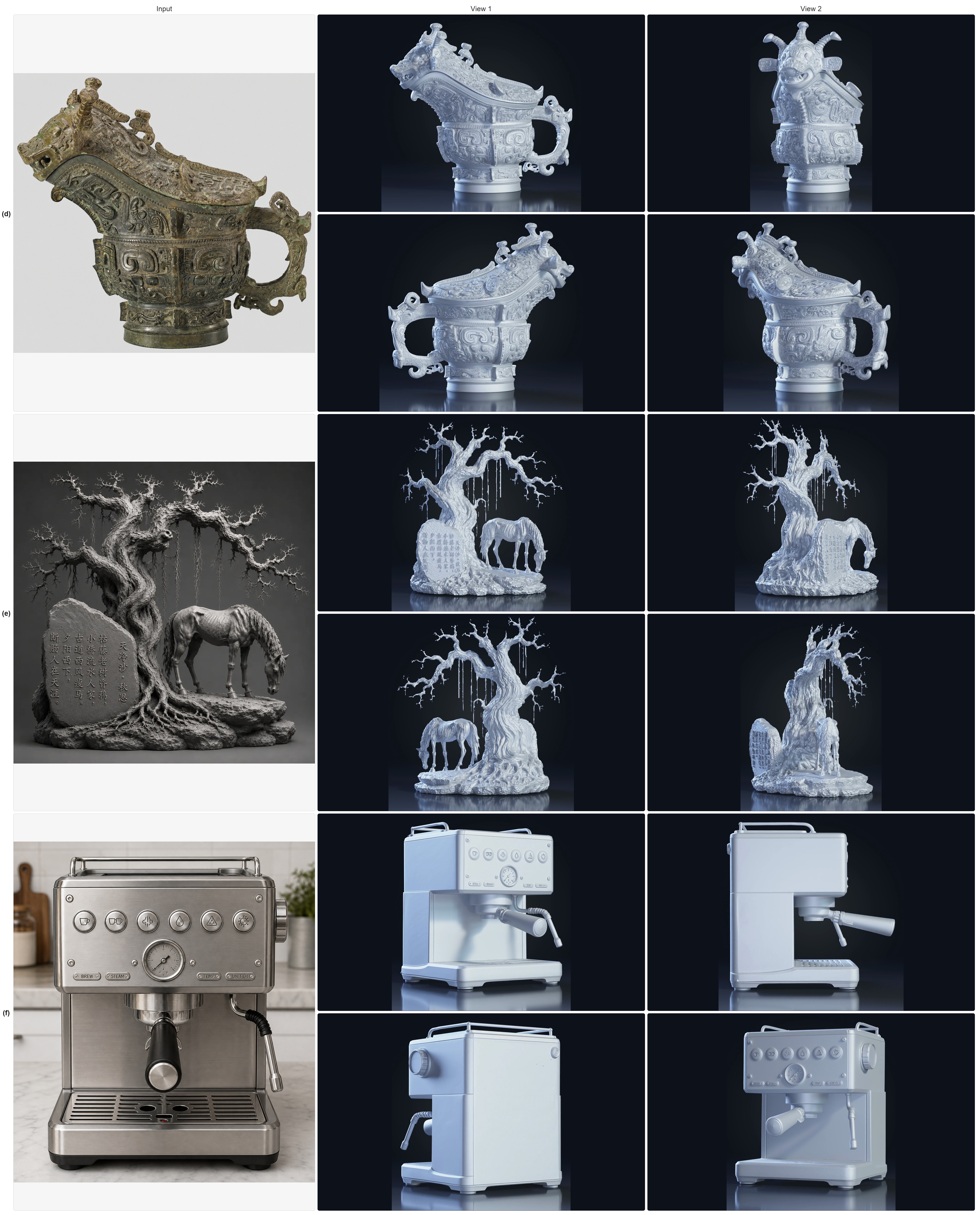}
    \caption{High-relief vessel; thin-structure scene; reflective mechanical object.}
    \label{fig:detailed-show-b}
  \end{subfigure}
  \caption{\textbf{\Tw{} results across six input categories.} In
  each case the left column is the input image and the $2\times 2$ block
  shows clay-shaded renderings of the generated mesh from complementary
  viewpoints.}
  \label{fig:detailed-show}
\end{figure}

Figure~\ref{fig:cmp-all} compares all five systems on five inputs, with two
regions of each enlarged. The five cases probe different failure modes.
Figure~\ref{fig:cmp-all}(a) probes scene completeness: Rodin Gen~2.5 discards
everything except the subject, and no detail metric registers this failure
because the detail it does produce is unobjectionable.
Figure~\ref{fig:cmp-all}(b) probes thin structure at close spacing.
Figure~\ref{fig:cmp-all}(c) probes fine relief. The medal carries lettering,
mid-scale relief, and fine rigging at once, so a system can succeed at one
scale and fail at another within a single result.
Figure~\ref{fig:cmp-all}(d) and~(e) probe facial identity with high-frequency
hair and fabric, where the input is a photograph of a known face.

Figures~\ref{fig:qual-compare-1} and~\ref{fig:qual-compare-2} probe the same
question at close range and at full render resolution. Each row shows one frame of a close-up camera path,
the same frame for every system, with two regions enlarged beneath it. The four
rows cover a winged figure, a gilded qilin, a cabinet of carved masks, and a fox
immortal statue. They test wing feathers and a crown band, the filigree band and
scroll of a medallion, an engraved name plate with the ornament beneath it, and a
lock of hair with a forehead mark. At this scale the systems agree on the shape
of the object and differ on the features that identify it.

Figures~\ref{fig:detailed-show-a} and~\ref{fig:detailed-show-b} show further
\Hi{} results across six input categories: high-detail busts, multi-object
scenes, full-body characters, high-relief vessels, thin-structure scenes,
and reflective mechanical objects. Each case shows the input alongside
clay-shaded renderings from complementary viewpoints.

\subsection{Summary}
\label{sec:eval:summary}

In sum, \Hi{} obtains the best value on all four automatic metrics and on
all five rate columns of the inscription benchmark. It is rated the same
or better in the majority of judgements in every pairwise human comparison.
We note three negative cases in the same results. First, at $0.757$ on
multi-view plausibility, close to a quarter of back-view completions fail a
criterion that asks only whether the surface was completed. Second, a
$16.5\%$ miss rate means roughly one character in six falls below
legibility, and seven characters were located but misread. The model omits
rather than invents, but the capability is not saturated. Last, $33$ inputs
support the group separation on detail alignment in Table~\ref{tab:main},
where the gap between the two groups is at least $0.087$. But they do not
support orderings within a group. The $0.0088$ between \Hi{} and Meshy~7 on
detail alignment, and the one point spanning four systems on aesthetics,
are within the range that per-input variance on a set this size can produce.

\FloatBarrier

\FloatBarrier
\bibliographystyle{plainnat}
\bibliography{references}

\appendix
\section{Contributors}
\label{sec:contributors}

\subsection*{Core Contributors}
Ziying Li, Shengchu Zhao, Huiang He, Yiyang Chen.

\subsection*{Contributors}
Listed alphabetically by last name: Jianwen Huang, Bailin Li, Changhao Li,
Jianhui Li, Jie Li, Ruiyang Liu, Yibo Luo, Tengjiao Sun, Pei Tang, Shiwen Wang,
Jiaqi Wu, Kang Wu, Kaiqiao Yang, Zherui Yang, Hu Zhang, Xuezhi Zhao,
Zihan Zhao, Xinhe Zheng.

\subsection*{Project Leader}
Yukun Li, Heliang Zheng, Rongfei Jia.

\end{document}